\documentclass{bmvc2k}

\title{Visual Tripwires: Anticipating Failure in Deep Vision Systems}

\usepackage{amsmath}
\usepackage{amssymb}
\usepackage{amsfonts}
\usepackage{amsthm}
\usepackage{multicol}
\usepackage{multirow}
\usepackage{booktabs}

\addauthor{Anoushka Harit}{}{1}
\addauthor{Rehan Zuberi}{}{1}
\addauthor{William Prew}{}{1}
\addauthor{Florian Markowetz}{}{1}

\addinstitution{
Cancer Research UK Cambridge Institute\\
University of Cambridge\\
Cambridge, UK
}

\runninghead{Student, Prof, Collaborator}{BMVC Author Guidelines}

\begin{document}

\maketitle

\begin{abstract}
Deep vision systems remain vulnerable to corruption, occlusion, and distribution shift despite strong benchmark performance. Existing reliability methods typically evaluate uncertainty at individual time steps and do not explicitly model how a system progresses toward failure. We introduce \textit{Visual Tripwires}, a predictive reliability framework that uses temporal instability in model behaviour to anticipate impending failure. Our central hypothesis is that predictive degradation develops progressively through measurable changes in latent representations, prediction trajectories, and attention structure. Visual Tripwires captures these changes using representation drift, prediction oscillation, trajectory curvature, and attention entropy. A lightweight tripwire predictor aggregates these signals over a temporal window to estimate the probability of failure within a future prediction horizon. Experiments across multiple datasets, architectures, and progressive perturbation settings show that the proposed instability signals emerge before predictive degradation and provide earlier and more accurate failure warnings than conventional uncertainty estimation methods. These results demonstrate that temporal instability contains useful information about future model reliability and provides a practical basis for early warning in deep vision systems.
\end{abstract}

%-------------------------------------------------------------------------

\section{Introduction}
\label{sec:intro}
Deep vision systems have achieved remarkable success across image classification, object detection, segmentation, tracking, and video understanding tasks \cite{he2016deep,dosovitskiy2020image,liu2021swin}. Despite strong benchmark performance, they remain highly vulnerable to corruption, occlusion, and distribution shift \cite{hendrycks2019benchmarking,geirhos2020shortcut,ovadia2019can}. In safety relevant settings such as content moderation or driving perception, this vulnerability is consequential: a model that fails silently can propagate an incorrect decision before a human has the chance to intervene.

Existing reliability methods, including calibration and out of distribution detection \cite{guo2017calibration,lakshminarayanan2017simple,hendrycks2016baseline}, underlie selective prediction, where a model defers to human review once confidence falls below a threshold. However, these methods assess reliability independently at each time step and typically detect unreliability only after degradation has occurred, leaving little lead time for review before a failure propagates.

Complex dynamical systems often exhibit early warning signals prior to critical transitions, with increasing variance and structural disruption preceding failure \cite{scheffer2009early}. Motivated by this, we treat failure in vision systems as a process rather than an isolated event, one that unfolds through instability in internal representation dynamics before it is visible in model outputs. This is supported by evidence that latent representations encode structural information about model behaviour \cite{raghu2017svcca,morcos2018insights} and are sensitive to distribution shift \cite{wang2020tent,sun2020test}, yet existing reliability methods rarely model this temporal evolution.

We introduce \textit{Visual Tripwires}, a framework that models the temporal evolution of latent representations, prediction trajectories, and attention structure to anticipate unreliable operating regimes before predictive collapse, providing lead time for human review. We define a family of instability signals, representation drift, prediction oscillation, latent trajectory curvature, and attention entropy, and aggregate them into a tripwire score that estimates the probability of future failure within a prediction horizon. This reframes reliability estimation as a temporal instability modeling problem, better suited to selective prediction under sustained perturbation.

We evaluate our framework across multiple architectures, datasets, and progressive perturbation settings, including a driving scene dataset relevant to deployment. Instability signals consistently rise before predictive degradation, providing substantially longer early warning than conventional uncertainty estimation. While our experiments focus on benign corruption, the same signature, sustained deviation building over time, is also expected under adaptive or adversarial perturbation, motivating temporal instability modeling as a direction for robust, verifiable vision systems.

\section{Related Work}
\label{sec:related}
Reliability in deep vision has primarily been addressed through per instance uncertainty estimation. Maximum softmax probability \cite{hendrycks2016baseline} and temperature scaling \cite{guo2017calibration} calibrate confidence for a single prediction; Monte Carlo dropout \cite{gal2016dropout} and deep ensembles \cite{lakshminarayanan2017simple} approximate uncertainty through sampling; ODIN \cite{liang2017enhancing} refines confidence using input perturbations. All condition only on the current prediction, so they detect unreliability only once it is already visible in the output. Selective prediction builds on these scores to decide when a model should abstain or defer to human review \cite{geifman2017selective,geifman2019selectivenet}, but inherits the same limitation: deferral is triggered at the point of failure, not before it. Our framework targets exactly this gap, providing the lead time that selective prediction requires to be preventive rather than reactive.

Vision models degrade sharply under corruption and distribution shift \cite{hendrycks2019benchmarking,geirhos2020shortcut}, and uncertainty estimates are known to remain poorly calibrated under exactly these conditions \cite{ovadia2019can}. Test time adaptation methods such as Tent \cite{wang2020tent} and test time training \cite{sun2020test} respond to this by updating model parameters online, treating representation instability as a fault to correct. We instead treat that instability as the signal itself, exploiting it to anticipate failure rather than to repair the model. This is grounded in evidence that latent representations carry structural information about generalization, typically probed through canonical correlation analysis \cite{raghu2017svcca,morcos2018insights}. Such analyses compare static snapshots of a trained network; we instead track the temporal evolution of a single network's representations along a trajectory, converting a diagnostic tool into an online monitoring signal.

Detecting distributional change in streaming data is a long standing problem in statistics, addressed by methods such as the Page-Hinkley test, CUSUM, and adaptive windowing \cite{gama2014survey}. These operate on scalar output statistics and are agnostic to model internals. Our instability signals are model internal analogues of the same principle, computed over latent representations and attention structure rather than output metrics, and combined through a learned aggregation rather than a fixed statistical test. The underlying hypothesis, that systems display measurable irregularity before a critical transition, is established in complex systems research, where rising variance and oscillation are known to precede collapse in ecological, climate, and biological systems \cite{scheffer2009early}. This principle has so far been applied to physical and biological dynamical systems; we apply it to the latent trajectory of a vision model under perturbation, treating predictive failure as the critical transition to be anticipated.

Video anomaly detection identifies frames or segments that deviate from normal appearance or motion \cite{liu2018future,sultani2018real}, but targets anomalies in the scene, not instability in the model perceiving it. Our framework asks a different question: not whether the input is anomalous, but whether the model's own response to it is becoming unstable, which we argue is the more direct signal of impending failure.

\section{Problem Formulation}
\label{sec:formulation}

Consider a temporally ordered sequence of visual inputs:

\begin{equation}
X = \{x_t\}_{t=1}^{T},
\qquad
x_t \in \mathbb{R}^{H \times W \times C},
\label{eq:input_sequence}
\end{equation}

where \(x_t\) denotes an image or video frame observed at time step \(t\). Given an input \(x_t\), a vision model \(f_\theta\) produces a predictive distribution over \(K\) task outcomes:

\begin{equation}
p_t = f_\theta(x_t) \in \Delta^{K-1},
\label{eq:model_prediction}
\end{equation}

where \(\Delta^{K-1}\) denotes the \(K\)-class probability simplex. We additionally extract an intermediate latent representation:

\begin{equation}
z_t = \phi_\theta(x_t) \in \mathbb{R}^{d}.
\label{eq:latent_representation}
\end{equation}

At time \(t\), the recent behaviour of the model is represented by a temporal window of length \(w\):

\begin{equation}
\mathcal{H}_t =
\left\{(z_\tau,p_\tau)\right\}_{\tau=t-w+1}^{t}.
\label{eq:temporal_history}
\end{equation}

This history captures temporal changes in both the internal representations and model predictions.

Let \(r_t\) denote a task-specific reliability score, and let \(\gamma\) be the minimum acceptable reliability threshold. We define a persistent failure state as:

\begin{equation}
F_t =
\mathbb{I}
\left[
r_\tau < \gamma,
\quad
\forall \tau \in \{t-m+1,\ldots,t\}
\right],
\label{eq:failure_state}
\end{equation}

where \(m\) is the number of consecutive time steps required to distinguish persistent degradation from an isolated prediction error.

Given a future prediction horizon \(\Delta\), the early warning target is defined as:

\begin{equation}
Y_t^{(\Delta)} =
\mathbb{I}
\left[
\max_{\tau \in \{t+1,\ldots,t+\Delta\}} F_\tau = 1
\right].
\label{eq:future_failure}
\end{equation}

Thus, \(Y_t^{(\Delta)}=1\) when the model enters a failure state at any point within the next \(\Delta\) time steps.

The objective is to learn a tripwire function \(g_\psi\) that estimates the probability of future failure from the recent model trajectory:

\begin{equation}
q_t =
g_\psi(\mathcal{H}_t)
\approx
P\left(
Y_t^{(\Delta)}=1
\mid
\mathcal{H}_t
\right).
\label{eq:tripwire_probability}
\end{equation}

A warning is activated when the estimated failure probability exceeds a decision threshold \(\eta\):

\begin{equation}
A_t = \mathbb{I}\left[q_t \geq \eta\right],
\label{eq:warning_activation}
\end{equation}

where \(A_t=1\) indicates an early warning. The threshold \(\eta\) is selected using validation data.

\begin{figure*}[htbp]
\centering
\includegraphics[width=0.75\textwidth]{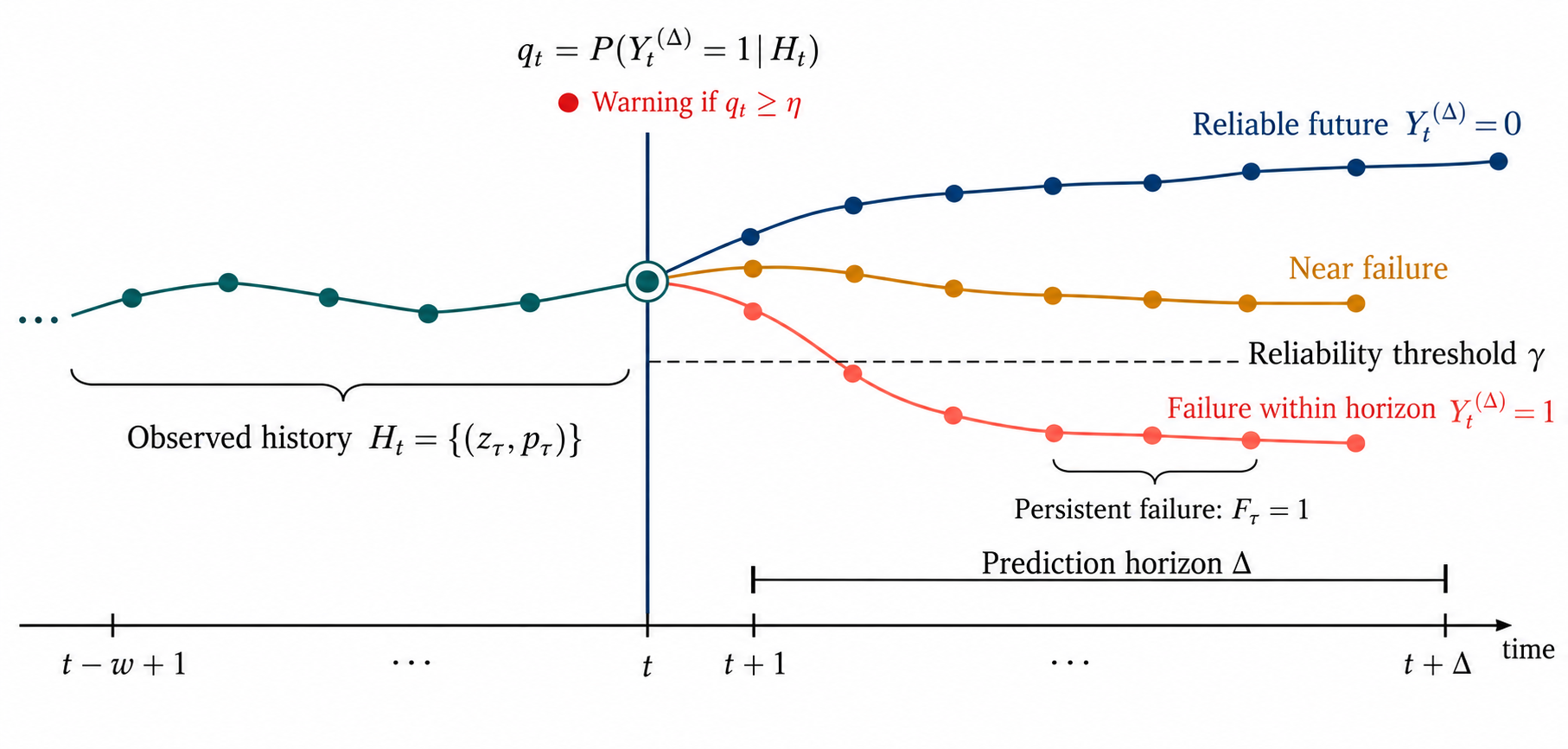}
\caption{Predictive failure formulation. Given the observed model history $\mathcal{H}_t$, the objective is to estimate the probability that the model enters a persistent failure state within the future horizon $\Delta$.}
\label{fig:problem_formulation}
\end{figure*}

We hypothesize that impending failure is preceded by measurable temporal instability. Stable operation produces smooth representation trajectories and consistent predictions, whereas emerging failure produces increasing representation drift, prediction oscillation, trajectory curvature, and attention uncertainty. Visual Tripwires uses these temporal signals to identify an emerging unreliable regime before the model crosses the predefined failure threshold.

\section{Method}
\label{sec:method}

\begin{figure*}[htbp]
\centering
\includegraphics[width=0.95\textwidth]{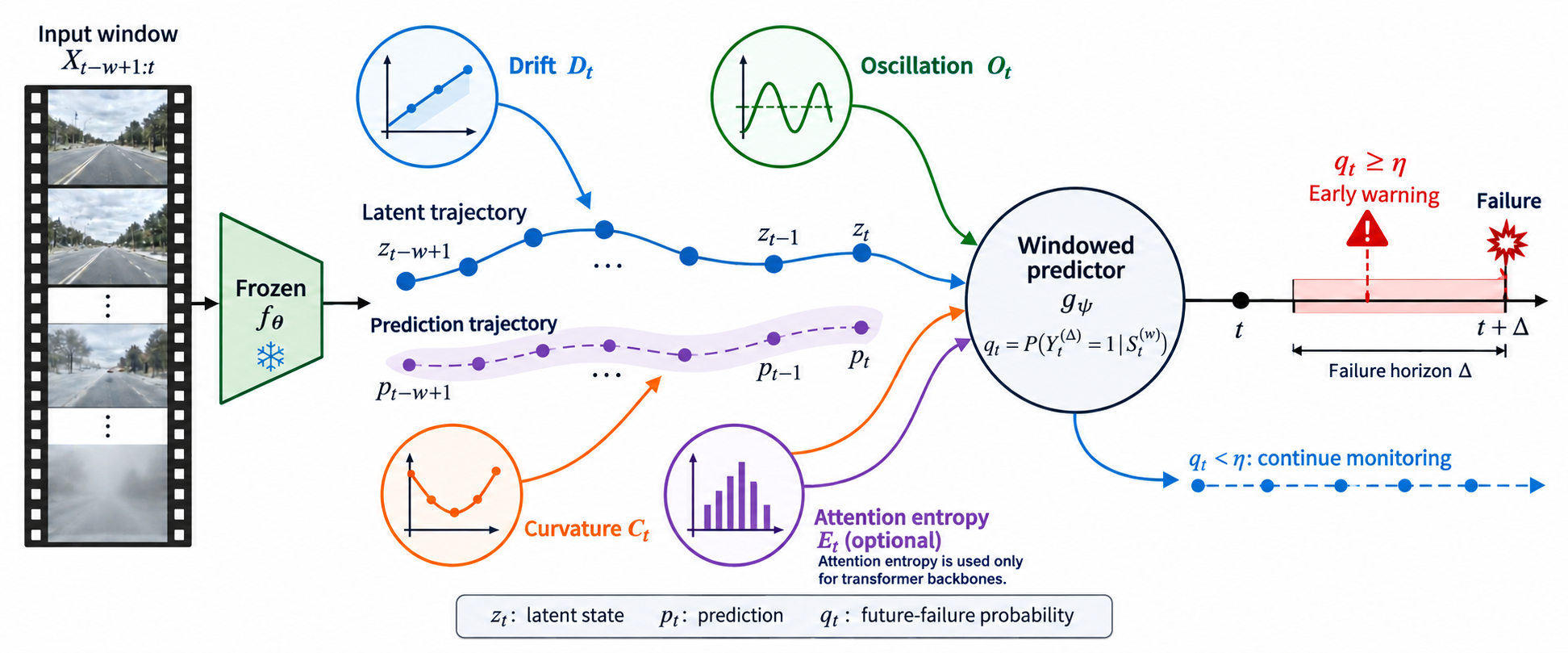}
\caption{Overview of Visual Tripwires. Instability signals derived from latent representations, predictions, and attention maps are aggregated over a temporal window to estimate the probability of future model failure.}
\label{fig:framework}
\end{figure*}

We introduce \textit{Visual Tripwires}, a framework that predicts impending model failure from the temporal evolution of internal representations and predictions. Given the model history \(\mathcal{H}_t\) defined in Section~\ref{sec:formulation}, we compute a set of instability signals, aggregate their recent values using a lightweight prediction network, and estimate the probability that the model will enter an unreliable state within the next \(\Delta\) time steps. Figure~\ref{fig:framework} presents an overview of the framework.

\subsection{Temporal Instability Signals}

Let \(z_t \in \mathbb{R}^{d}\) and \(p_t \in \Delta^{K-1}\) denote the latent representation and predictive distribution produced at time \(t\). We compute four complementary instability signals. All signals are defined for \(t \geq 3\), allowing the first and second temporal differences to be evaluated.

\paragraph{Representation Drift.}
Representation drift measures the displacement between consecutive latent states:

\begin{equation}
D_t = \left\|z_t-z_{t-1}\right\|_2.
\label{eq:representation_drift}
\end{equation}

Large values indicate rapid changes in the internal representation of temporally adjacent inputs.

\paragraph{Prediction Oscillation.}
Prediction oscillation measures the change in the model's predictive distribution:

\begin{equation}
O_t = \left\|p_t-p_{t-1}\right\|_2.
\label{eq:prediction_oscillation}
\end{equation}

This signal increases when the model produces inconsistent predictions across consecutive observations.

\paragraph{Trajectory Curvature.}
We approximate the local curvature of the latent trajectory using its discrete second difference:

\begin{equation}
C_t =
\left\|z_t-2z_{t-1}+z_{t-2}\right\|_2.
\label{eq:trajectory_curvature}
\end{equation}

Unlike representation drift, which measures displacement, curvature captures changes in the direction and rate of movement through the representation space.

\paragraph{Attention Entropy.}
For transformer architectures, let \(A_t^{h}(i,j)\) denote the attention assigned by query token \(i\) to key token \(j\) in head \(h\), where

\begin{equation}
\sum_{j=1}^{n} A_t^{h}(i,j)=1.
\label{eq:attention_normalization}
\end{equation}

Attention entropy is defined as:

\begin{equation}
E_t =
-\frac{1}{Hn}
\sum_{h=1}^{H}
\sum_{i=1}^{n}
\sum_{j=1}^{n}
A_t^{h}(i,j)
\log\left(A_t^{h}(i,j)+\epsilon\right),
\label{eq:attention_entropy}
\end{equation}

where \(H\) is the number of attention heads, \(n\) is the number of tokens, and \(\epsilon>0\) ensures numerical stability. Higher entropy indicates more diffuse attention. For convolutional architectures, the attention signal is omitted and the tripwire predictor operates on the remaining three signals.

\subsection{Signal Normalization}

Because the instability signals have different numerical scales, each signal is standardized using statistics estimated from clean reference sequences. For a signal \(S_t \in \{D_t,O_t,C_t,E_t\}\), we compute:

\begin{equation}
\widetilde{S}_t =
\frac{S_t-\mu_S}{\sigma_S+\epsilon},
\label{eq:signal_normalization}
\end{equation}

where \(\mu_S\) and \(\sigma_S\) are the mean and standard deviation of \(S\) on the clean reference split. These statistics remain fixed during validation and testing.

The normalized instability vector is:

\begin{equation}
s_t =
\begin{bmatrix}
\widetilde{D}_t &
\widetilde{O}_t &
\widetilde{C}_t &
\widetilde{E}_t
\end{bmatrix}^{\top}.
\label{eq:instability_vector}
\end{equation}

For convolutional architectures, \(s_t\) contains only \(\widetilde{D}_t\), \(\widetilde{O}_t\), and \(\widetilde{C}_t\).

\subsection{Visual Tripwire Predictor}

To distinguish sustained instability from isolated fluctuations, we concatenate the instability vectors from a trailing window of length \(w\):

\begin{equation}
S_t^{(w)} =
\operatorname{vec}
\left(
s_{t-w+1},\ldots,s_t
\right).
\label{eq:windowed_instability}
\end{equation}

A lightweight multilayer perceptron \(g_\psi\) maps the windowed signals to a scalar logit:

\begin{equation}
a_t = g_\psi\left(S_t^{(w)}\right).
\label{eq:tripwire_logit}
\end{equation}

The Visual Tripwire score is the corresponding probability of failure within the future horizon:

\begin{equation}
q_t =
\sigma(a_t)
\approx
P\left(
Y_t^{(\Delta)}=1
\mid
S_t^{(w)}
\right),
\label{eq:tripwire_score}
\end{equation}

where \(\sigma(\cdot)\) is the sigmoid function and \(Y_t^{(\Delta)}\) is the future failure label defined in Equation~\ref{eq:future_failure}. A warning is issued when:

\begin{equation}
q_t \geq \eta,
\label{eq:tripwire_warning}
\end{equation}

where \(\eta\) is selected on the validation set. The base vision model \(f_\theta\) remains fixed, and only the parameters \(\psi\) of the tripwire predictor are optimized.

\subsection{Training Objective}

The tripwire predictor is trained using binary cross entropy over all valid temporal windows:

\begin{equation}
\mathcal{L}_{\mathrm{BCE}}
=
-\frac{1}{N}
\sum_{t \in \mathcal{I}}
\left[
Y_t^{(\Delta)}\log q_t
+
\left(1-Y_t^{(\Delta)}\right)
\log\left(1-q_t\right)
\right],
\label{eq:bce_loss}
\end{equation}

where \(\mathcal{I}\) is the set of valid training indices and \(N=|\mathcal{I}|\).

We additionally introduce a temporal regularization term to discourage unstable changes in consecutive tripwire scores:

\begin{equation}
\mathcal{L}_{\mathrm{temp}}
=
\frac{1}{N-1}
\sum_{t \in \mathcal{I}}
\left(q_t-q_{t-1}\right)^2.
\label{eq:temporal_loss}
\end{equation}

The complete training objective is:

\begin{equation}
\mathcal{L}
=
\mathcal{L}_{\mathrm{BCE}}
+
\lambda_{\mathrm{temp}}
\mathcal{L}_{\mathrm{temp}},
\label{eq:total_loss}
\end{equation}

where \(\lambda_{\mathrm{temp}}\) controls the strength of temporal regularization. The binary cross entropy term trains the model to predict future failure, while the temporal term reduces isolated fluctuations in the warning score.

\section{Experimental Setup}
\label{sec:setup}

We evaluate \textit{Visual Tripwires} for predictive failure estimation under progressive corruption and distribution shift. Given a temporally perturbed input sequence, the objective is to predict whether the model will enter an unreliable regime before degradation becomes observable in its outputs.

\subsection{Datasets}

We evaluate on CIFAR-10-C and ImageNet-C \cite{hendrycks2019benchmarking}, which extend CIFAR-10 \cite{krizhevsky2009learning} and ImageNet \cite{deng2009imagenet} with 15 corruption types at five severity levels. Because these benchmarks do not provide dense temporal sequences, we construct a trajectory for each clean image by interpolating corruption intensity across and beyond the benchmark severity levels. This produces a sequence \(X=\{x_1,\ldots,x_T\}\) with monotonically increasing severity. For discrete corruptions, such as JPEG compression, each severity level is held for \(T/5\) consecutive steps. Trajectories are generated independently for each corruption type, and results are averaged across corruptions unless stated otherwise.

We additionally evaluate on BDD100K \cite{yu2020bdd100k}, which provides native temporal structure with variation in illumination, motion, and weather. Perturbations are applied along the existing temporal axis, providing a complementary evaluation under realistic changes in visual content.

Training, validation, and test partitions are separated at the source-image or video-clip level to prevent correlated frames from appearing across splits. A subset of corruption types is excluded from training and evaluated separately to measure generalization to unseen distribution shifts.

\subsection{Architectures}

We evaluate ResNet-50 \cite{he2016deep}, ViT-B/16 \cite{dosovitskiy2020image}, and Swin-T \cite{liu2021swin}. Latent representations are extracted from the penultimate feature layer of ResNet-50 and the final CLS token embeddings of ViT-B/16 and Swin-T. For the transformer models, attention entropy is computed across all heads in the final layer. For ResNet-50, attention entropy is omitted and \(g(\cdot)\) is trained using the remaining three instability signals.

\subsection{Progressive Failure Protocol}

We evaluate Gaussian noise, motion blur, defocus blur, brightness variation, contrast degradation, JPEG compression, and structured occlusion. Perturbation severity increases monotonically with time, producing gradual degradation in visual quality and representation stability.

For classification tasks, failure is triggered when accuracy remains below \(50\%\) for five consecutive time steps. For BDD100K, failure is triggered when the predicted class changes more than three times within a five-step window or when maximum softmax confidence falls below \(0.4\). All thresholds are selected on the validation split and fixed during testing.

\subsection{Baselines}

We compare Visual Tripwires with Maximum Softmax Probability \cite{hendrycks2016baseline}, Predictive Entropy \cite{malinin2018predictive}, Temperature Scaling \cite{guo2017calibration}, Monte Carlo Dropout \cite{gal2016dropout}, Deep Ensembles \cite{lakshminarayanan2017simple}, and ODIN \cite{liang2017enhancing}. These baselines evaluate reliability independently at each time step, whereas Visual Tripwires models instability over a trailing temporal window.

\subsection{Evaluation Metrics}

We report AUROC, AUPRC, warning lead time, and false alarm rate on held-out test sequences. Lead time measures the number of steps between tripwire activation and failure and is reported only for sequences that contain a failure event. False alarm rate is the proportion of non-failing sequences in which the tripwire activates.

Each experiment is repeated over five runs with different random seeds and a fixed data split. Where available, results are reported as mean \(\pm\) standard deviation; otherwise, the mean across the five runs is reported. We use inverse-frequency class weighting in the binary cross-entropy loss to account for the imbalance between stable and future-failure windows.

\subsection{Implementation Details}

All experiments are implemented in PyTorch using publicly available pretrained checkpoints without finetuning on corrupted data. We use a sliding window of \(w=5\), a prediction horizon of \(\Delta=10\), and a sequence length of \(T=100\), unless stated otherwise. The values of \(w\) and \(\Delta\) are selected through grid search on the validation split.

The aggregation function \(g(\cdot)\) is trained using Adam with a learning rate of \(10^{-4}\) and a batch size of \(32\). The loss weights \(\lambda_{\mathrm{pred}}\) and \(\lambda_{\mathrm{temp}}\), together with the decision threshold \(\tau\), are selected on the validation split and fixed during testing. Experiments are conducted on NVIDIA 2080 Ti GPUs.

\section{Results}
We evaluated visual tripwires across multiple architectures, datasets, and perturbation settings to assess whether temporal instability dynamics provides reliable early warning signals prior to predictive collapse. We compare our framework against standard uncertainty estimation and reliability baselines under progressive corruption and distribution shift.

\subsection{Failure Prediction Across Datasets} 
We first evaluate failure prediction performance across multiple datasets and vision architectures. Table~\ref{tab:main_results} reports AUROC and lead time under progressive perturbations. Visual Tripwires consistently outperform conventional uncertainty estimation methods across all datasets and architectures. In addition to achieving the highest AUROC, the proposed framework provides substantially larger lead time prior to failure events, demonstrating that instability in latent dynamics emerges before predictive degradation becomes observable.

\begin{table*}[htbp]
\centering
\small
\caption{Failure prediction performance across datasets and architectures.}
\label{tab:main_results}
\begin{tabular}{llcccc}
\toprule
Dataset & Method & ResNet50 & ViT-B & Swin-T & Lead Time \\
\midrule

\multirow{6}{*}{CIFAR-10-C}
& MSP & 0.71 $\pm$ 0.01 & 0.69 $\pm$ 0.02 & 0.72 $\pm$ 0.01 & 2.1 \\
& Predictive Entropy & 0.73 $\pm$ 0.01 & 0.72 $\pm$ 0.01 & 0.74 $\pm$ 0.01 & 2.5 \\
& MC Dropout & 0.76 $\pm$ 0.01 & 0.75 $\pm$ 0.01 & 0.77 $\pm$ 0.01 & 3.0 \\
& Deep Ensembles & 0.79 $\pm$ 0.01 & 0.78 $\pm$ 0.01 & 0.80 $\pm$ 0.01 & 3.4 \\
& ODIN & 0.81 $\pm$ 0.01 & 0.80 $\pm$ 0.01 & 0.82 $\pm$ 0.01 & 3.6 \\
& Visual Tripwires (Ours) & \textbf{0.88 $\pm$ 0.01} & \textbf{0.89 $\pm$ 0.01} & \textbf{0.90 $\pm$ 0.01} & \textbf{6.7} \\

\midrule

\multirow{6}{*}{ImageNet-C}
& MSP & 0.68 $\pm$ 0.02 & 0.66 $\pm$ 0.02 & 0.69 $\pm$ 0.01 & 1.9 \\
& Predictive Entropy & 0.71 $\pm$ 0.01 & 0.70 $\pm$ 0.01 & 0.72 $\pm$ 0.01 & 2.3 \\
& MC Dropout & 0.74 $\pm$ 0.01 & 0.73 $\pm$ 0.01 & 0.75 $\pm$ 0.01 & 2.9 \\
& Deep Ensembles & 0.78 $\pm$ 0.01 & 0.77 $\pm$ 0.01 & 0.79 $\pm$ 0.01 & 3.2 \\
& ODIN & 0.80 $\pm$ 0.01 & 0.79 $\pm$ 0.01 & 0.81 $\pm$ 0.01 & 3.5 \\
& Visual Tripwires (Ours) & \textbf{0.87 $\pm$ 0.01} & \textbf{0.88 $\pm$ 0.01} & \textbf{0.89 $\pm$ 0.01} & \textbf{6.4} \\

\midrule

\multirow{6}{*}{BDD100K}
& MSP & 0.69 $\pm$ 0.01 & 0.68 $\pm$ 0.01 & 0.70 $\pm$ 0.01 & 2.0 \\
& Predictive Entropy & 0.72 $\pm$ 0.01 & 0.71 $\pm$ 0.01 & 0.73 $\pm$ 0.01 & 2.4 \\
& MC Dropout & 0.75 $\pm$ 0.01 & 0.74 $\pm$ 0.01 & 0.76 $\pm$ 0.01 & 3.1 \\
& Deep Ensembles & 0.79 $\pm$ 0.01 & 0.78 $\pm$ 0.01 & 0.80 $\pm$ 0.01 & 3.5 \\
& ODIN & 0.81 $\pm$ 0.01 & 0.80 $\pm$ 0.01 & 0.82 $\pm$ 0.01 & 3.7 \\
& Visual Tripwires (Ours) & \textbf{0.89 $\pm$ 0.01} & \textbf{0.90 $\pm$ 0.01} & \textbf{0.91 $\pm$ 0.01} & \textbf{7.1} \\

\bottomrule
\end{tabular}
\end{table*}

\subsection{Temporal Instability Emerges Before Failure}
\begin{figure*}[t]
    \centering
    \includegraphics[width=0.90\textwidth]{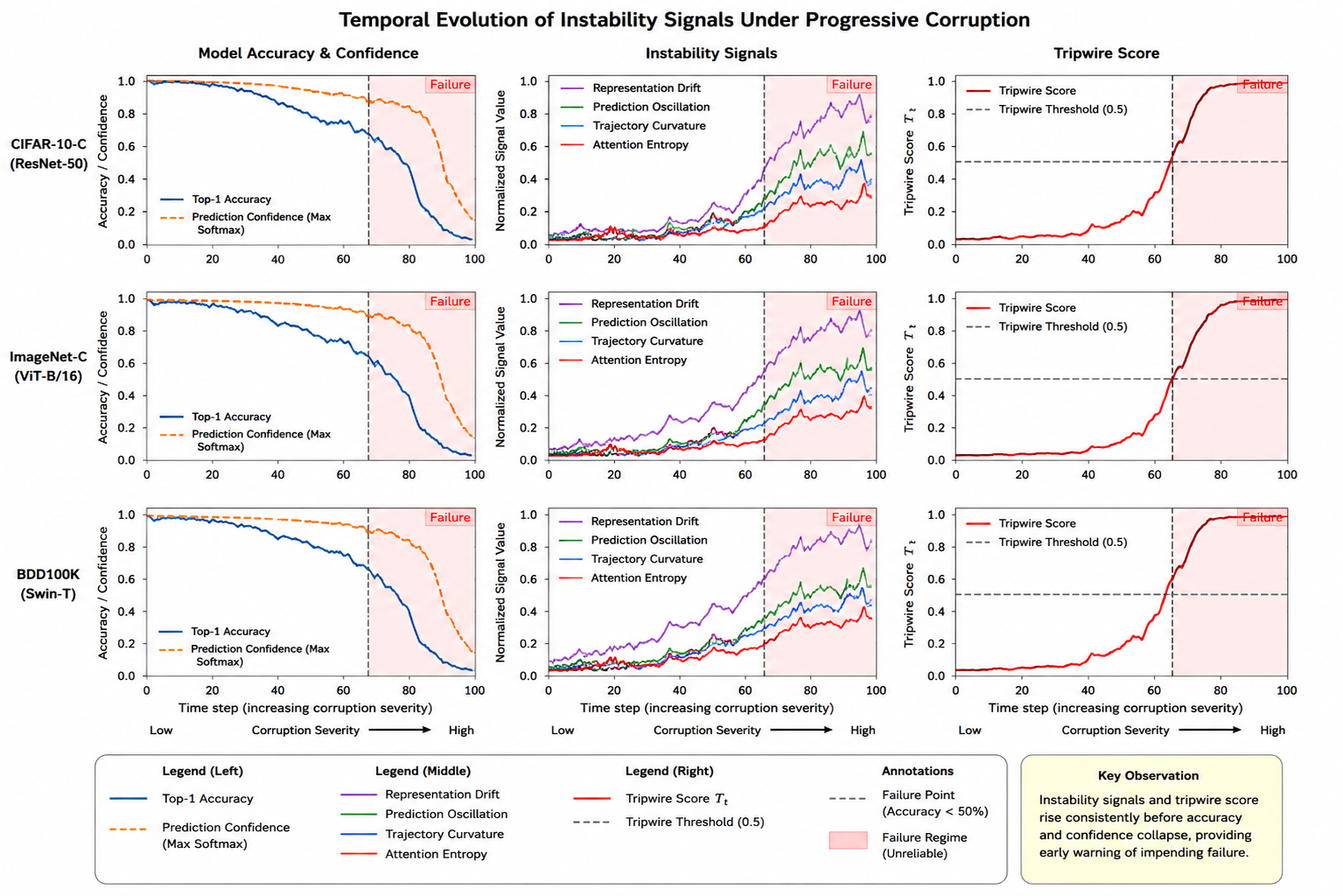}
    \caption{
    Temporal evolution of instability signals under progressive corruption. Instability dynamics and tripwire scores rise consistently before predictive degradation, providing early warning signals of impending failure.
    }
    \label{fig:tripwire_dynamics}
\end{figure*}

We next analyze the temporal evolution of instability signals under progressive perturbations. Figure~\ref{fig:tripwire_dynamics} visualizes representation drift, prediction oscillation, and tripwire scores together with model accuracy.

Across all datasets, the instability signals increase before predictive degradation. In particular, prediction confidence often remains high while latent instability increases substantially, indicating that representation dynamics contain predictive information that is not captured by conventional uncertainty measures. We also observe increasing trajectory fragmentation as the severity of the perturbation increases, suggesting that predictive degradation is preceded by structural instability in the latent feature space.

\subsection{Case Study: Early Detection of Failure Under Progressive Occlusion}
\begin{figure*}[htbp]
    \centering
    \includegraphics[width=\textwidth]{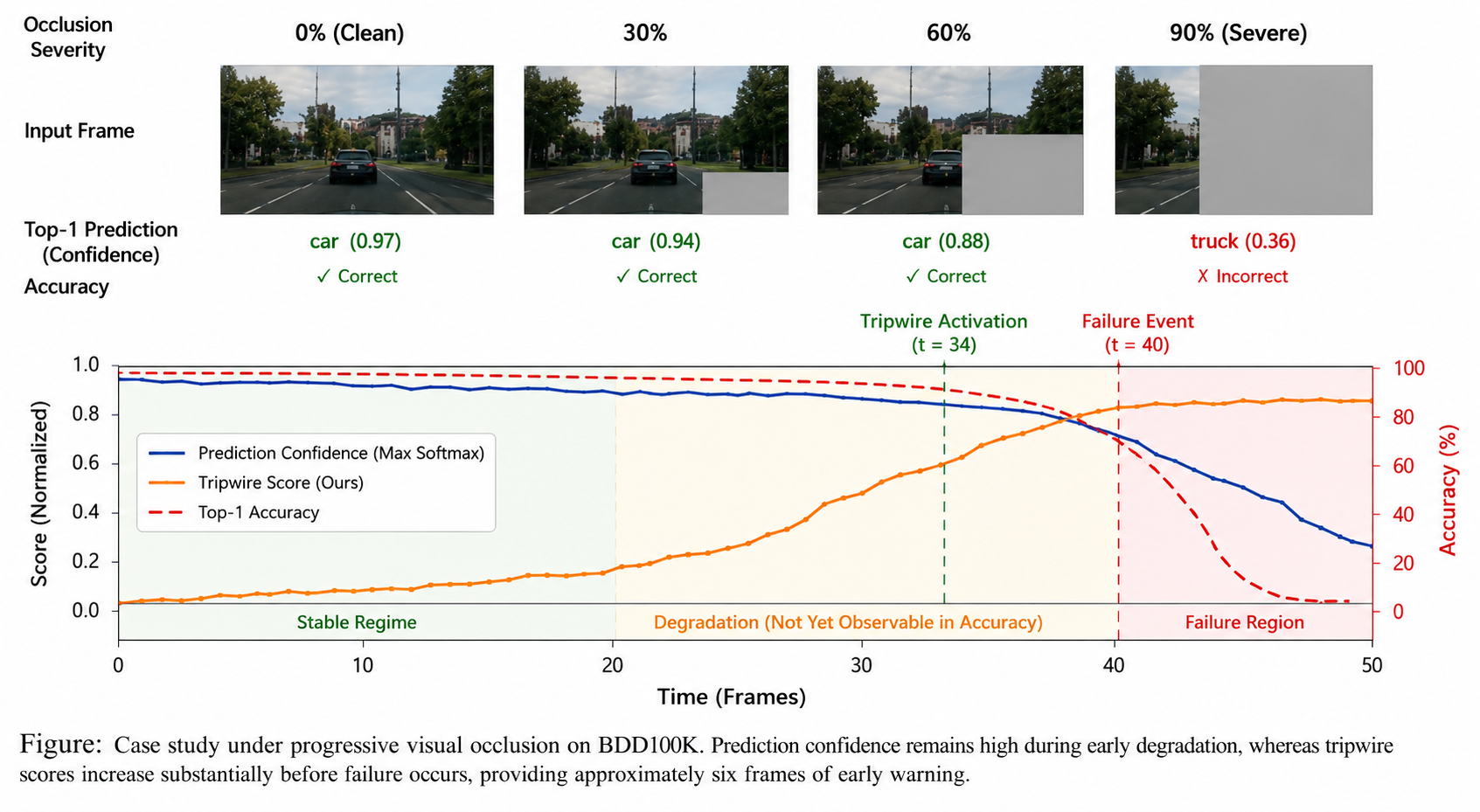}
    \caption{
    Case study under progressive visual occlusion. Prediction confidence remains high during the early stages of degradation, whereas tripwire scores increase substantially before predictive failure becomes observable.
    }
    \label{fig:case_study}
\end{figure*}

To better understand the behaviour of Visual Tripwires in realistic settings, we analyze a representative failure trajectory from BDD100K under progressive visual occlusion. Figure~\ref{fig:case_study} illustrates the evolution of prediction confidence, tripwire score, and model accuracy as occlusion severity increases over time.

As occlusion severity increases, the model initially maintains high prediction confidence despite progressive degradation in latent representation stability. In contrast, the tripwire score rises steadily and activates approximately six time steps before the failure event.

Notably, confidence based reliability measures remain largely unchanged until shortly before prediction accuracy deteriorates, whereas instability signals provide a substantially earlier indication of degradation. This behaviour suggests that latent representation dynamics contain predictive information about future failure that is not captured by conventional confidence estimates.

These observations provide qualitative evidence for our central hypothesis that predictive failure emerges progressively through instability in representation geometry rather than as an instantaneous prediction error.

\subsection{Robustness Under Distribution Shift}
We further evaluate Visual Tripwires under unseen corruption types and out of distribution settings. Table~\ref{tab:ood_results} reports failure prediction performance under unseen perturbations.

\begin{table}[htbp]
\centering
\caption{Failure prediction under unseen distribution shifts.}
\label{tab:ood_results}
\begin{tabular}{lcc}
\toprule
Method & AUROC $\uparrow$ & False Alarm Rate $\downarrow$ \\
\midrule
MSP & 0.65 & 0.31 \\
Entropy & 0.69 & 0.28 \\
MC Dropout & 0.73 & 0.24 \\
Ensemble & 0.76 & 0.21 \\
ODIN & 0.79 & 0.18 \\
\midrule
Ours & \textbf{0.86} & \textbf{0.12} \\
\bottomrule
\end{tabular}
\end{table}

Visual Tripwires consistently outperform uncertainty based baselines under unseen perturbations, achieving higher AUROC and lower false alarm rates. While conventional confidence measures often become unreliable under strong corruption, instability signals extracted from latent trajectories remain sensitive to structural degradation, enabling earlier and more robust failure detection.

\subsection{Ablation Study}
We evaluate the contribution of each instability signal using a leave-one-component-out ablation. Each configuration is evaluated over five independent runs, and Table~\ref{tab:ablation} reports the mean failure prediction AUROC and mean warning lead time across these runs.

\begin{table}[htbp]
\centering
\small
\caption{Ablation of individual instability signals. Results are averaged over five independent runs.}
\label{tab:ablation}
\resizebox{\columnwidth}{!}{
\begin{tabular}{lcc}
\toprule
Configuration & Mean AUROC $\uparrow$ & Mean Lead Time (steps) $\uparrow$ \\
\midrule
Without Representation Drift & 0.82 & 5.0 \\
Without Prediction Oscillation & 0.84 & 5.3 \\
Without Trajectory Curvature & 0.85 & 5.6 \\
Without Attention Entropy & 0.86 & 5.8 \\
\midrule
Full Model & \textbf{0.90} & \textbf{7.1} \\
\bottomrule
\end{tabular}
}
\end{table}

The full model achieves the highest mean AUROC and the longest mean warning lead time. Removing representation drift causes the largest performance reduction, decreasing AUROC by 0.08 and lead time by 2.1 steps. Removing prediction oscillation produces the second-largest reduction, followed by trajectory curvature and attention entropy. Performance declines whenever a component is removed, indicating that the four instability signals provide complementary information for anticipating predictive failure. Representation drift and prediction oscillation make the largest observed contributions under this evaluation.

\subsection{Qualitative Analysis}
\begin{figure*}[htbp]
    \centering
    \includegraphics[width=0.75\textwidth]{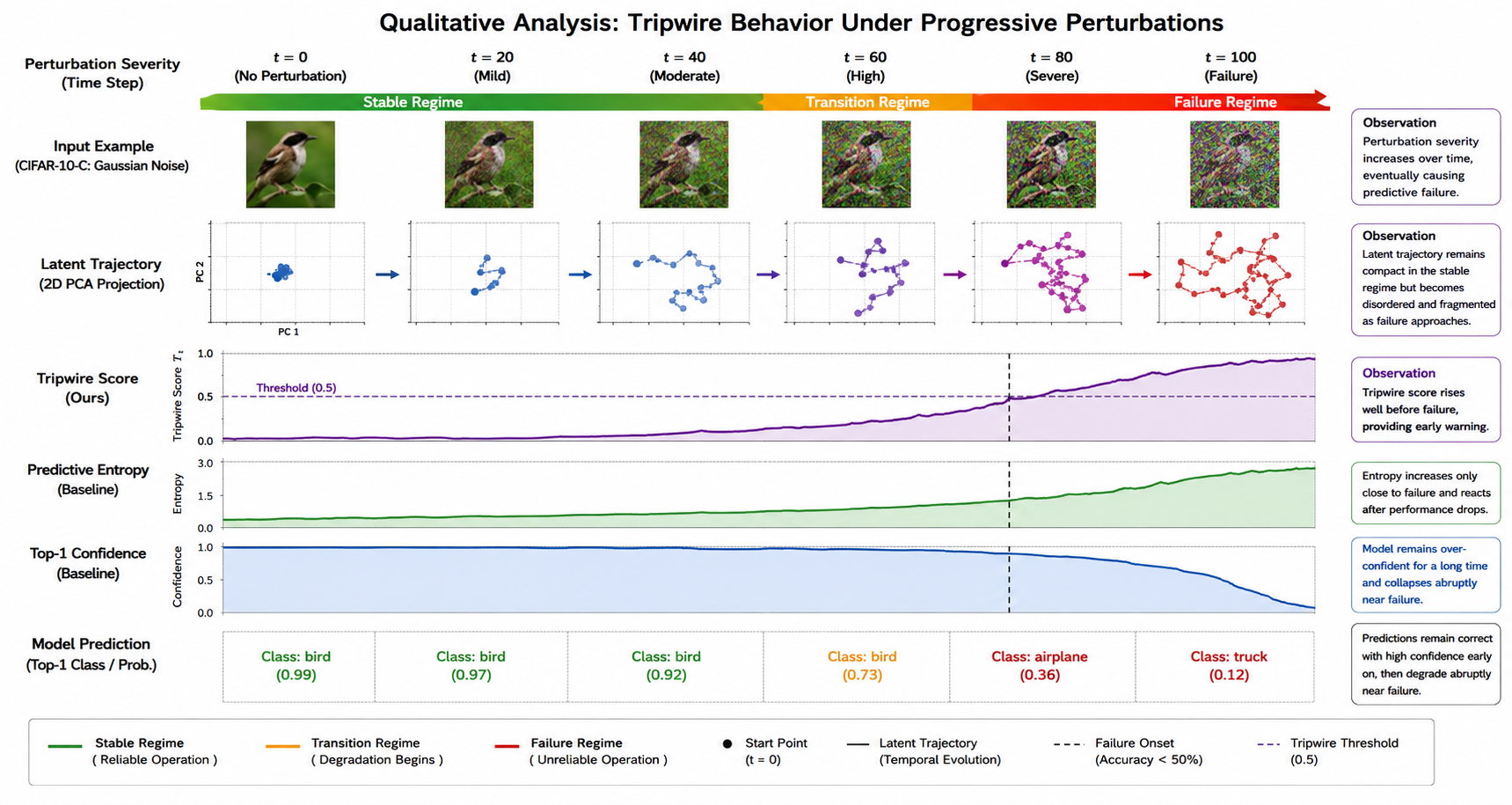}
    \caption{
    Qualitative analysis of Visual Tripwires under progressive perturbations. Stable operating regimes exhibit smooth latent trajectories and low tripwire scores, while impending failure is characterized by trajectory fragmentation and increasing instability signals.
    }
    \label{fig:qualitative_results}
\end{figure*}

Figure~\ref{fig:qualitative_results} provides a qualitative visualization of representation dynamics under progressive perturbations.

Stable operating regimes are associated with smooth and coherent latent trajectories. As perturbation severity increases, representation trajectories become increasingly fragmented, accompanied by rising tripwire scores. These qualitative observations are consistent with the quantitative results and support the hypothesis that predictive degradation is preceded by structural changes in latent representation geometry.

\section{Conclusion}
We introduced \textit{Visual Tripwires}, a predictive reliability framework that uses temporal changes in model behaviour to identify impending failures in deep vision systems. The framework combines representation drift, prediction oscillation, trajectory curvature, and attention entropy to detect emerging instability before a predefined failure event.

Experiments across CIFAR-10-C, ImageNet-C, and BDD100K demonstrate that Visual Tripwires consistently outperform confidence and uncertainty based baselines in failure prediction AUROC while providing longer warning lead times. The framework also remains effective under unseen distribution shifts, achieving higher detection performance and fewer false alarms. The ablation results show that each instability signal contributes to performance, with representation drift and prediction oscillation producing the largest observed gains. Together, these findings indicate that temporal changes in latent representations and predictions provide information about impending failure that is not fully captured by individual confidence estimates.

Future work will evaluate Visual Tripwires on dense prediction tasks, multimodal models, and naturally occurring distribution shifts. A further direction is to connect tripwire activation to adaptive responses, such as abstention, human review, input restoration, or dynamic model selection, enabling systems to act on early warnings before reliability deteriorates.

\subsection{Limitations}
Visual Tripwires require a temporally ordered sequence of inputs and may therefore be less suitable for isolated predictions without meaningful temporal or perturbation structure. Warning reliability may also decrease under abrupt domain transitions, where failure occurs before a sufficiently long instability trajectory can be observed, or in highly stochastic environments, where normal variation may resemble failure related instability.

The framework requires access to intermediate representations and, for attention based signals, internal attention maps. This limits direct application to closed source systems that expose only output probabilities or predicted labels. The current instability measures are manually specified and may not capture all forms of model degradation. Their thresholds may also require calibration when transferring the framework across architectures, datasets, or deployment conditions.

Finally, the evaluation primarily uses controlled corruptions and progressive perturbations. Although these settings enable precise measurement of failure onset and warning lead time, they do not represent the full complexity of deployment environments, where multiple shifts may occur simultaneously and failure boundaries may be ambiguous. Evaluation on naturally occurring temporal shifts and prospective deployment data is therefore required to establish practical generalization.

\bibliography{egbib}

\end{document}